\pdfoutput=1

\documentclass[11pt]{article}

\usepackage[]{EMNLP2022}
\usepackage{times}
\usepackage{latexsym}

\usepackage{amsmath}
\usepackage{amssymb}
\usepackage{bm}
\usepackage{booktabs}
\usepackage{graphicx}
\usepackage{multicol}
\usepackage{multirow}

\usepackage[T1]{fontenc}

\usepackage[utf8]{inputenc}
\usepackage{pmboxdraw}

\usepackage{microtype}

\usepackage{inconsolata}

\newcommand{\supth}{$^\text{th}$ }

\title{SpeechNet: Weakly Supervised, End-to-End Speech \\Recognition at Industrial Scale}

\newcommand{\parheader}[1]{{\bf \smallskip \noindent #1.}}

\DeclareMathOperator{\LF}{LF}

\author{Raphael Tang,$^1$ Karun Kumar,$^1$ Gefei Yang,$^1$ Akshat Pandey,$^1$ Yajie Mao,$^1$  \\
{\bf Vladislav Belyaev,$^1$ Madhuri Emmadi,$^1$ Craig Murray,$^1$ Ferhan Ture,$^1$ Jimmy Lin$^2$} \vspace{1mm}\\
$^1$Comcast Applied AI \hspace{5mm}$^2$University of Waterloo\\
{\small $^1$\texttt{firstname\_lastname@comcast.com} \hspace{5mm} $^2$\texttt{jimmylin@uwaterloo.ca}}}

\begin{document}
\maketitle
\begin{abstract}
End-to-end automatic speech recognition systems represent the state of the art, but they rely on thousands of hours of manually annotated speech for training, as well as heavyweight computation for inference.
\hspace{-0.5mm}Of course, this impedes commercialization since most companies lack vast human and computational resources.
\hspace{-0.5mm}In this paper, we explore training and deploying an ASR system in the label-scarce, compute-limited setting.
\hspace{-0.5mm}To reduce human labor, we use a third-party ASR system as a weak supervision source, supplemented with labeling functions derived from implicit user feedback.
\hspace{-0.5mm}To accelerate inference, we propose to route production-time queries across a pool of CUDA graphs of varying input lengths, the distribution of which best matches the traffic's.
Compared to our third-party ASR, we achieve a relative improvement in word-error rate of 8\% and a speedup of 600\%. 
\hspace{-0.5mm}Our system, called SpeechNet, currently serves 12 million queries per day on our voice-enabled smart television.
\hspace{-0.5mm}To our knowledge, this is the first time a large-scale, Wav2vec-based deployment has been described in the academic literature.

\end{abstract}

\section{Introduction}

Training an end-to-end automatic speech recognition (ASR) model requires hundreds, if not thousands, of hours of hand-labeled speech.
With the rise of silicon-hungry pretrained transformers, these models additionally need increasing amounts of computational power just to perform inference.
Together, these two hurdles impede effective model deployment at all but the largest technology companies and specialized speech processing startups.
The hurdles certainly apply to us at Comcast, the main stage of this work.
Our industrial challenge is to fine-tune and deploy a large, pretrained speech recognition model, without an army of annotators (as in Amazon) or mammoth GPU farms (e.g., Google).
Our end application is the Xfinity X1, a voice-enabled smart television serving millions of active devices in the United States.

Evidently, cloud ASR services are cheaply available.\footnote{But not cheaper or better than using our own in-house ASR system; otherwise, there would be no need for this work!}
Google Cloud, for example, charges \$1.44 USD per hour of transcribed speech.
In contrast, manual annotation services like Rev cost \$90 \textit{per hour}, and our in-house annotators, whom Comcast must use to protect user privacy, cost even more.
Thus, cloud ASR's comparatively low pricing, combined with its decent quality, suggests its utility as an annotation source in the absence of substantial human-labeled data.

Nevertheless, cloud ASR still falls short of human parity and hence demands label denoising. 
To do this, we propose to use implicit user feedback to remove incorrectly labeled examples, bootstrapping an existing cloud ASR service.
We derive these labeling functions using signals from query repetition, session length, and ASR confidence scores.
We model them in Snorkel~\cite{ratner2017snorkel}, a popular data programming framework, producing a 1400-hour weakly labeled dataset.
Trained on this, our models improve over those using unfiltered data by an average 0.97 points in word-error rate (WER), as presented in Section \ref{sec:results}.

As for the second hurdle of resource efficiency, many model acceleration methods exist.
However, few meet our productionization criteria: we seek to preserve the quality, ruling out structured pruning~\cite{li2020efficient}; we wish to preserve the pretrained architectural structure, eliminating knowledge distillation~\cite{tang2019distilling}; and we require stable software--GPU support, disqualifying low bit-width quantization~\cite{shen2020q} and other CPU-oriented approaches.

All things considered, the prime candidates are medium bit-width quantization, decoder optimizations~\cite{abdou2004beam}, and CUDA computation graphs~\cite{gray2019getting}.
The first two follow the literature, but the third is more open ended.
In spite of their record-breaking performance, CUDA graphs work only with fixed-length input, not variable length.
Toward this, we propose to allocate a pool of CUDA graphs of varying lengths, altogether matching the production-time traffic length distribution.
During inference, we route each query to the graph with the least upper-bound in length.
As we show in Section \ref{sec:results}, this yields a 3--5$\times$ increase in throughput.


We claim the following contributions: first, we derive novel labeling functions for constructing weakly labeled speech datasets from in-production ASR systems, improving our best model by a relative 8\% in word-error rate.
Second, we propose to accelerate model inference using a pool of CUDA graphs, attaining a 7--9$\times$ inference speed increase at no quality loss.
The resulting system, SpeechNet, currently serves more than 20 million queries per day on our smart television.
To our knowledge, we are the first to describe a large-scale, Wav2vec-based deployment in the academic literature.

\section{Our SpeechNet Approach}

Our task is to train and deploy a state-of-the-art, end-to-end ASR system, without using human-annotated data.
The context of this deployment is a smart TV, which users interact with using a speech-driven remote control.
To issue a voice query, users hold a button, speak their command, and release the button.
We initially serve them with a third-party cloud ASR service, bootstrapping it for the development of SpeechNet.
Data-wise, we store thousands of hours of utterances per day, complete with session IDs, transcripts, and device IDs.
Resource-wise, we have 30 deployment nodes, each hosting an Nvidia Tesla T4 GPU and receiving 120 queries per second (QPS) at peak time; thus, our model's real-time factor must exceed 120.

\subsection{End-to-End ASR Modeling}
In end-to-end ASR systems, we transcribe speech waveform directly to orthography, consolidating the traditional acoustic--pronunciation--language modeling approach.
Similar to natural language processing, the dominant paradigm in speech is to pretrain transformers~\cite{vaswani2017attention} on unlabeled speech using an unsupervised contrastive objective, then fine-tune on labeled datasets~\cite{baevski2020wav2vec}.
We practitioners further fine-tune these released models on our in-domain datasets.

Concretely, we feed an audio amplitude sequence $(x_t)_{t=1}^\ell \in [-1, 1]$ into a pretrained model consisting of one-dimensional convolutional feature extractors and transformer layers, getting frame-level context vectors $(\bm h_t)_{t=1}^N \in \mathbb{R}^k$.
On each of these vectors, we perform a softmax transformation across the vocabulary $V$, for a final probability distribution sequence of $(\bm y_t)_{t=1}^N \in \mathbb{R}^{|V|}$.
For fine-tuning, we use a training set composed of audio--transcript pairs and optimize with the standard connectionist temporal classification objective (CTC; \citealp{graves2012connectionist}) for speech recognition.
We uncase the transcripts and encode them with a character-based tokenizer, as is standard.
At inference time, we decode the CTC outputs with beam search and a four-gram language model.

\subsection{Data Curation}
\label{sec:lfs}
\begin{figure}
    \centering
    \includegraphics[scale=0.35]{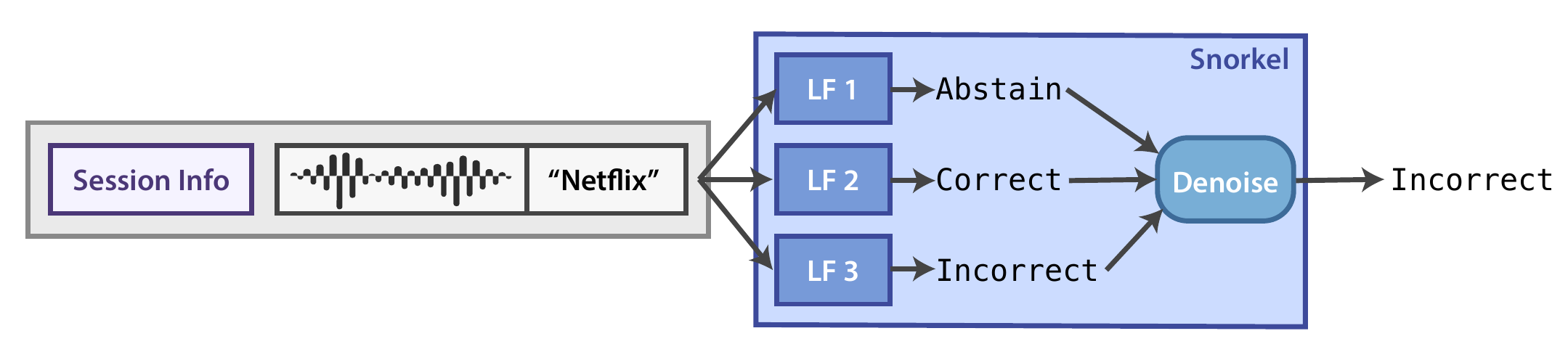}
    \caption{An example weak labeling. In this case, we would discard the incorrect transcript, ``Netflix.''}
    \label{fig:lfs}
\end{figure}

To build a weakly labeled dataset, we turn to Snorkel~\cite{ratner2017snorkel}, a popular data programming framework for aggregating and denoising weak labelers.
In Snorkel, domain experts first create handwritten weak labelers, which the authors call labeling functions (LFs).
Each of these LFs takes as input an unlabeled example, as well as any auxiliary data, and either outputs a label or abstains.
Next, Snorkel applies these LFs to each example in a dataset, producing a matrix of noisy labels.
It learns from this noisy observation matrix a generative model with the true labels as latent variables, which it supplies to downstream tasks.


Our task is to remove incorrect transcripts from a weakly constructed dataset.
Our LF inputs are audio clips and transcripts, along with session data, and our outputs are one of correct, incorrect, or abstain.
After Snorkel denoises the LF outputs and labels each dataset example, we discard abstained or incorrect ones, as visualized in Figure~\ref{fig:lfs}.
We derive and use the three following novel LFs:

\parheader{Session position}
We group queries in the same session if each occurs within 60 seconds of at least one other and is issued by the same user.
Previously, we found a negative correlation between the intrasession position of a query and the word-error rate~\cite{tang2019yelling}, where the last query consistently has a low word-error rate (WER), and long sessions have high intermediate query WERs.
With this finding, we write the session position LF, given query $q$, as

\vspace{-3mm}
\small
$$
\LF_\text{SP}(q) := \begin{cases}
\texttt{CORRECT} & \text{if $q$ is last in its session}\\
\texttt{INCORRECT} & \text{if sess. length $\geq 3$, $q$ not last}\\
\texttt{ABSTAIN} & \text{otherwise.}
\end{cases}
$$
\normalsize

\parheader{ASR confidence}
For each transcribed utterance, ASR systems output a confidence score, which correlates with the WER.
In most systems, this score results from an addition between the acoustic model score and the language model score.
The first is a function of speech, while the second of text.
Since our third-party ASR service is opaque, we have access only to the final score.
This complicates its direct use because thresholding it would skew the balance toward frequent words, as influenced by the language model.

To bypass this issue, we collect sample statistics of the final score \textit{grouped by transcript text}, then design an LF with transcript-specific thresholds.
This way, we remove the language model score as a confounder.
Define

\vspace{-3mm}
\small
$$
    \LF_\text{AC}(q) := \begin{cases}
        \texttt{CORRECT} & \text{if $s(q) \geq \text{p80}(q)$ }\\
        \texttt{INCORRECT} & \text{if $s(q) \leq \text{p20}(q)$}\\
        \texttt{ABSTAIN} & \text{if p20$(q)$ or p80$(q)$ undefined}\vspace{-1mm}\\
        & \text{or otherwise,}
    \end{cases}
$$
\normalsize
where $s(q)$ is the confidence score for query $q$ from the third-party ASR, and $\text{p20}(q)$ and $\text{p80}(q)$ return the 20\supth and 80\supth percentile ASR score for the transcript of $q$, respectively.
\begin{figure}[t]
    \centering
    \includegraphics[scale=0.24]{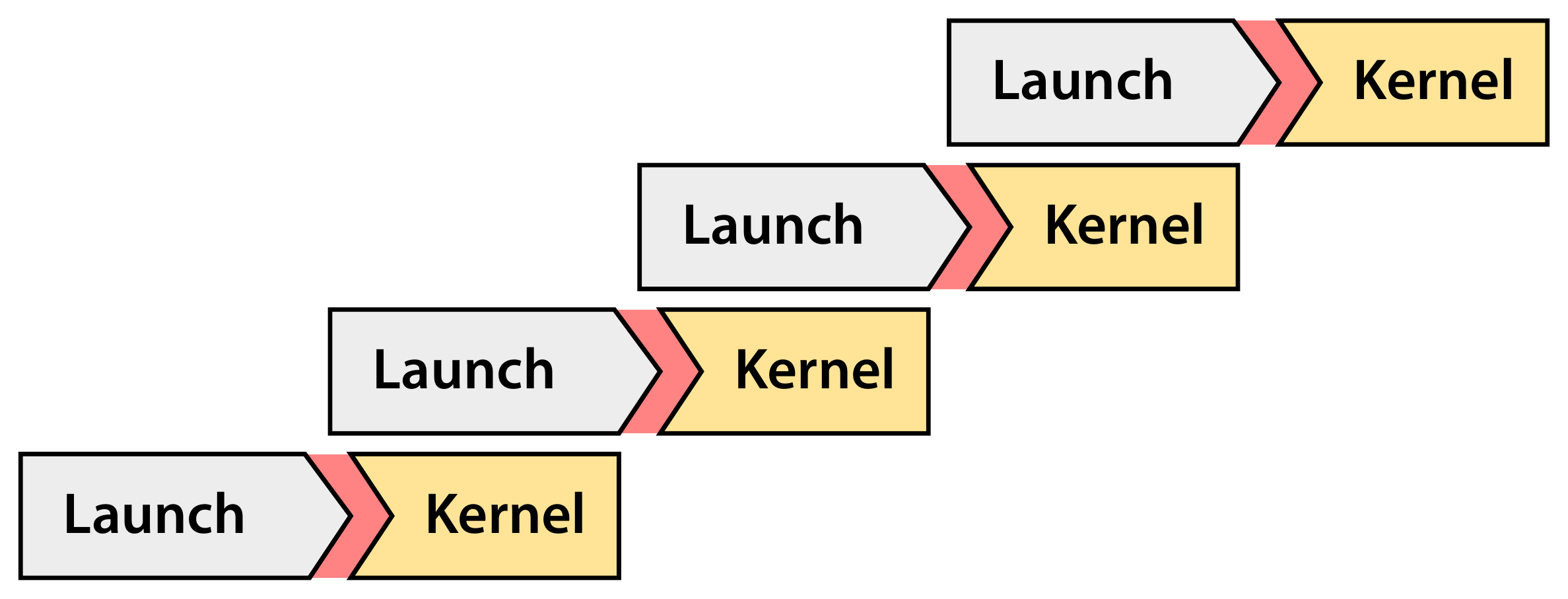}
    \caption{Typical way for the CPU to launch a sequence of small GPU kernels, with time flowing from left to right. Red area denotes launch latency.}\vspace{3mm}
    \label{fig:graph1}
    \includegraphics[scale=0.24,trim={1.85cm 0 0 0}]{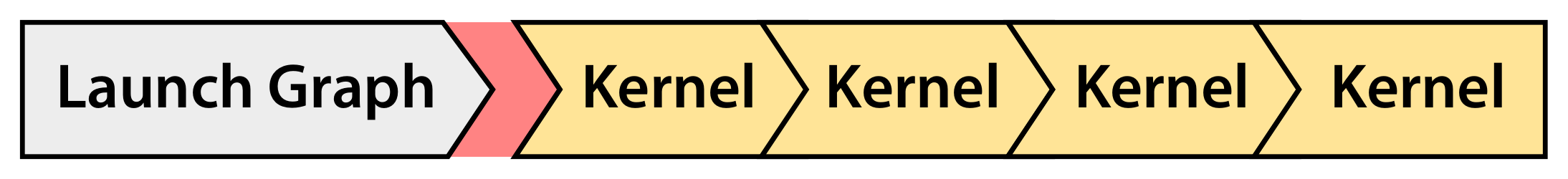}
    \caption{Launching a CUDA graph. Difference in right margin relative to Figure~\ref{fig:graph1} portrays time savings.}
    \label{fig:graph2}
\end{figure}

\parheader{Rapid repetition}
Users often rapidly repeat their voice queries upon ASR mistranscription~\cite{li2020auto}.
Given this, we can discard queries that closely precede others from the same user:

\vspace{-3mm}
\small
$$
    \LF_\text{RR}(q) := \begin{cases}
        \texttt{INCORRECT} & \text{if the user's next query}\vspace{-1mm} \\
        & \text{occurs $\leq$ 13 seconds of $q$}\\
        \texttt{ABSTAIN} & \text{otherwise.}
    \end{cases}
$$
\normalsize
On our platform, we've determined 13 seconds to be the optimal duration in terms of specificity and sensitivity~\cite{li2020auto}.


\subsection{Model Inference Acceleration}
\label{sec:graphs}

In production, we use a batch size of one for inference.
This largely decreases efficiency because GPU kernel launches now dominate the processing time, as portrayed in Figure~\ref{fig:graph1}.
In our case, we can't just pad to a large fixed size, since computation increases quadratically with length for transformers.
It's also infeasible to use batching (e.g., batch together sequential queries) because only 4--6 queries arrive in a 50-millisecond window per server, and we can't afford to sacrifice that much speed.

To improve inference efficiency, CUDA graphs allow a sequence of GPU kernels to be captured and run as a single computation graph, thus incurring one CPU launch operation instead of many---see Figure~\ref{fig:graph2}.
However, these graphs are input shape and control flow static, so they must be preconstructed.
This clearly poses a barrier to using variable-length audio as input.

\begin{figure}[t]
    \centering
    \includegraphics[scale=0.525]{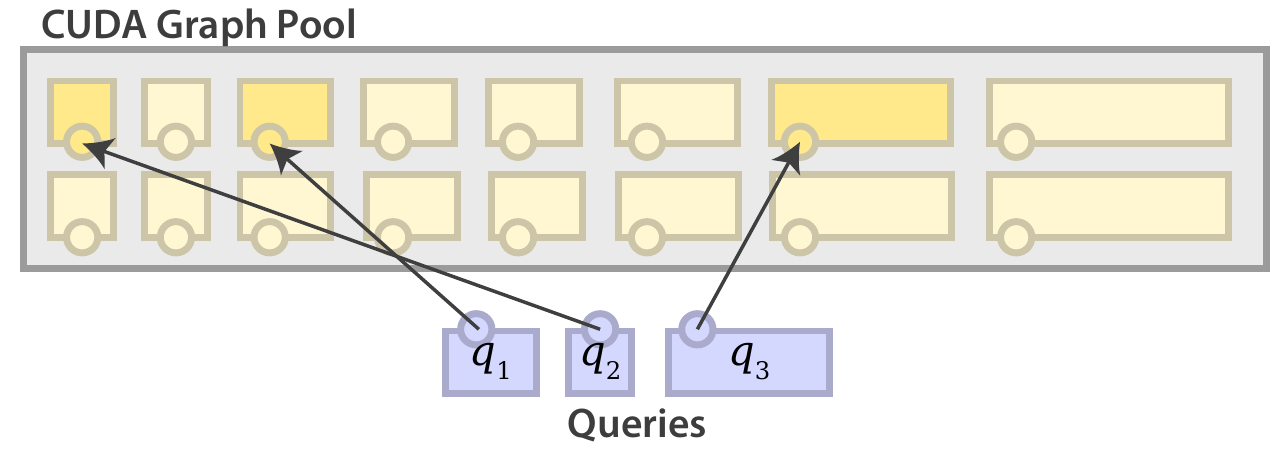}
    \caption{Three queries routed across a graph pool.}
    \label{fig:graph-pool}
\end{figure}
To address this issue, we propose to allocate a pool of differently sized CUDA graphs, then route each query to the nearest upper-bound graph.
For higher efficiency, we match the length distribution of the pool with that of the computation time on production traffic.
Formally, let $X$ be the random variable (r.v.) denoting the arrival distribution of the lengths of production-time queries.
Let $Z := f(X)$ be the time it takes for a CUDA graph to perform inference for length $X$.
Then, our CUDA graph pool comprises $\mathcal{G} := (g_{z_1}, \dots, g_{z_n})$, where $g_{z_i}$ denotes a CUDA graph of length $z_i$ and $z_1, \dots, z_n$ are realizations of $Z$.
To serve a query of length $l$, we pick the graph $g_{z^*}$, where
\begin{equation}
z^* := \min \{z_i ~|~ g_{z_i} \in \mathcal{G}, z_i \geq l\}.
\end{equation}
Our upstream system sends no more than ten seconds of audio by design, bounding this set.
We illustrate this process in Figure~\ref{fig:graph-pool}.





\section{Experimental Setup}

Our key experiments are to validate the model effectiveness of our labeling functions (Section \ref{sec:lfs}) and the computational savings of our CUDA graph pool (Section \ref{sec:graphs}).
We trained every run on one \texttt{p3.2xlarge} Amazon Web Services (AWS) instance, which has an Nvidia V100 GPU and eight virtual CPU cores.
We implemented our models in PyTorch using the HuggingFace Transformers library~\cite{wolf2019huggingface} and Nvidia's NeMo~\cite{kuchaiev2019nemo}; see the appendix for more details.

\begin{table}[t]
    \small
    \setlength{\tabcolsep}{2pt}
    \centering
    \begin{tabular}{lccc}
    \toprule[1pt]
         \bf Dataset & \bf Train/Dev/Test Hrs. & \bf \# Speakers & \bf \# Unique  \\\midrule
         CC-20 & 22/2.2/2.2 & 40K/4K/4K & 20 \\
         CC-LG & 1400/1.0/2.5 & 325K/2K/4K & 88K \\
    \bottomrule[1pt]
    \end{tabular}
    \caption{Dataset statistics. Further query distribution details are in the appendix.}
    \label{table:datasets}
\end{table}

\subsection{Dataset Curation}
We curated two datasets: one critical dataset, called CC-20, comprising the twenty most frequent commands, and another large-scale dataset, named CC-LG, consisting of audio examples sampled uniformly at random from user traffic.
We split our datasets into one or more training sets, a development (dev) set, and a test set, all drawn from separate days and speakers---see Table~\ref{table:datasets} for statistics.
On CC-20, native English speakers annotated the training set to establish an ``upper bound'' in quality, relative to using the weakly labeled datasets.
On CC-LG, the 1400-hour set was too large to annotate, so we skipped that.
On both datasets, we manually annotated the dev and test sets to serve as gold evaluation sets.

For the weakly labeled training sets, we constructed one set with raw transcripts from the third-party ASR system and another set with transcripts from Snorkel, filtered using the labeling functions in Section \ref{sec:lfs}.
We name the former set ``raw'' and the latter ``weak.''
To remove dataset size as a confounder, we use the same size for all training sets.

\subsection{Baselines and Models}
For our first baseline, we picked Google Cloud's public ASR offering~\cite{beaufays2022google}, primarily to sanity check our third-party ASR service.
We used their standard model offering, touted as state of the art, costing us \$0.006 USD per 15 seconds of speech. 
For our second baseline, we selected our third-party ASR service that we licensed from a major American technology company.

\begin{table}[t]
\setlength{\tabcolsep}{2.75pt}
\centering
\begin{tabular}{llcc}
\toprule[1pt]
     \multirow{2}{*}{\bf Model} & \bf \multirow{2}{*}{Training} & \bf CC-20 & \bf CC-LG \\
     & & Dev/Test & Dev/Test\\
     \midrule
     Google Cloud & \hspace{3mm}-- & 24.7/24.7 & 26.5/25.5 \\
     Our Third Party & \hspace{3mm}-- & 7.56/7.60 & 10.8/9.66 \\
     \midrule
     \multicolumn{4}{c}{Our Trained Models}\\
     \midrule
     \multirow{2}{*}{SEW$_\text{tiny}$} & Raw & 6.72/6.82 & 17.4/16.3 \\
     \multirow{2}{*}{\small 41M parameters} & Weak & 5.17/4.80 & 15.9/14.5 \\
     & Human & 4.79/4.66 & --\\
     \midrule
     \multirow{2}{*}{Wav2vec2.0$_\text{base}$} & Raw & 2.81/3.17 & 10.2/9.11 \\
     \multirow{2}{*}{\small 94M parameters} & Weak & 1.62/1.77 & \bf 9.14/8.82 \\
     & Human & \bf 1.54/1.75 & --\\
     \midrule
     \multirow{2}{*}{Conformer$_\text{large}$} & Raw & 3.52/3.68 & 12.6/10.6 \\
     \multirow{2}{*}{\small 120M parameters} & Weak & 3.63/4.08 & 12.0/9.78 \\
     & Human & 2.60/2.72 & --\\
\bottomrule[1pt]
\end{tabular}
\caption{Dev and test WERs of models trained on sets without LFs (raw), with LFs (weak), and with human annotations (human). Best results bolded.}
\label{table:main-results}
\end{table}
\begin{figure}[t]
    \centering
    \includegraphics[scale=0.43,trim={2mm 0 2.7mm 0},clip]{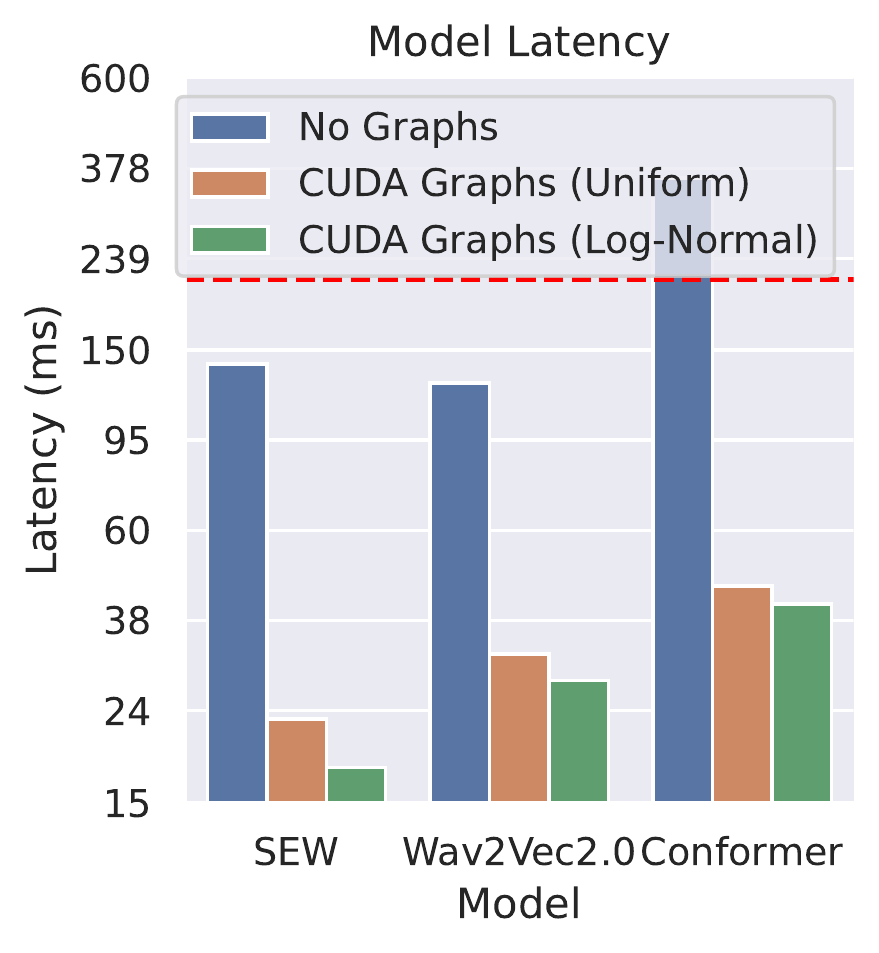}
    \includegraphics[scale=0.43,trim={2mm 0 2.7mm 0},clip]{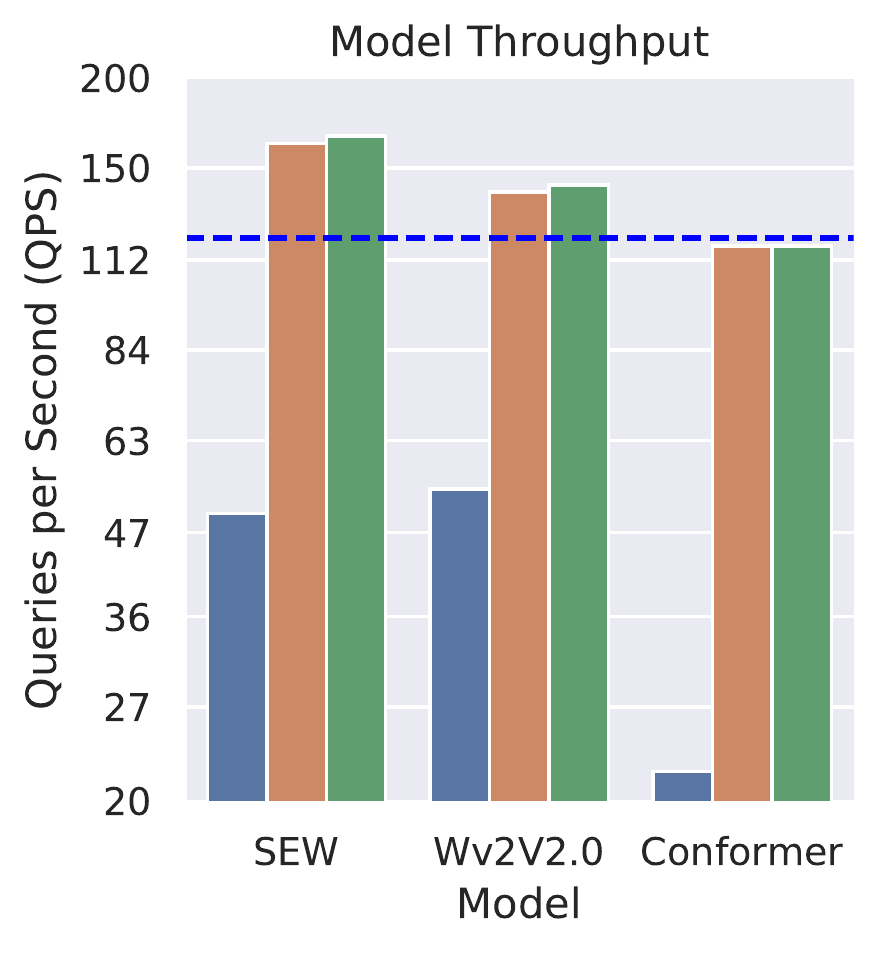}
    \caption{Throughput in queries per second and latency in milliseconds of all three models, under different CUDA graph pool settings. The red line on the left is our third-party ASR model latency and the blue line on the right our required throughput in production.}
    \label{fig:main-accel}
\end{figure}

\parheader{Models}
We chose three different state-of-the-art, pretrained transformer models from the literature, each representing a separate computational operating point: the Squeezed and Efficient Wav2vec model, tiny variant (SEW-tiny; \citealp{wu2022performance}), at 41 million parameters; the standard Wav2vec 2.0 base model (Wav2vec 2.0-base; \citealp{baevski2020wav2vec}), at 94 million parameters; and the large Conformer model (Conformer-large; \citealp{gulati2020conformer}), at 120 million.
We initialized them with LibriSpeech-fine-tuned weights and trained them using standard gradient-based optimization---we put details in the appendix.

\section{Results and Discussion}
\label{sec:results}

We present our model quality results in Table~\ref{table:main-results}.
Unsurprisingly, Google Cloud does worse than our third-party service, which has been specifically tailored to our in-domain vocabulary.
On average, sets curated with Snorkel (denoted as ``weak'') improves the WER by 0.97 points (95\% CI, 0.09 to 1.85) relative to those without (``raw'').
Wav2vec 2.0-base, our best model, outperforms the third party by a relative 70\% and 8\% on CC-20 and CC-LG, respectively.
Except for Conformer-large, all models trained on Snorkel-labeled sets achieve near parity with those on human-annotated training sets, with Wav2vec 2.0-base in particular reaching a test WER on CC-20 worse by only 0.02 points (1.77 vs. 1.75).
We speculate that conformers perform worse than Wav2vec 2.0-base does due to using log-Mel spectrograms instead of raw audio waveform: our voice queries greatly differ in loudness, resulting in exponential fluctuations after applying the log transform (as the input approaches 0).


\begin{table}[t]
\centering
\begin{tabular}{lcc}
\toprule[1pt]
     \bf \multirow{2}{*}{Training Set} & \bf CC-20 & \bf CC-LG \\
     & Dev/Test & Dev/Test\\
     \midrule
     Raw (no LFs) & 2.81/3.17 & 14.9/13.6 \\
     ~+ LF$_\text{SP}$ & 2.32/2.64 & 13.3/12.1 \\
     ~~+ LF$_\text{AC}$ & 2.16/1.93 & 13.3/11.9 \\
     ~~~+ LF$_\text{RR}$ & 1.62/1.77 & 13.1/11.8 \\
     \midrule
     Human & 1.52/1.75 & --\\
\bottomrule[1pt]
\end{tabular}
\caption{Quality of Wav2vec 2.0-base under differently constructed but equally sized training sets.}
\label{table:abl-model}
\end{table}

We chart our model acceleration results in Figure~\ref{fig:main-accel}.
We gather these statistics from replaying production-time traffic as fast as possible to saturate the model.
Overall, CUDA graph pools accelerate our models by 7--9$\times$ (left subfigure; compare blue and green bars) and increase throughput by 3--5$\times$ (right subplot).
Initializing the graph lengths to be log-normal distributed ekes out a few percentage points (compare orange and green) in performance, since that better matches our production traffic.
Most stark is the contrast between vanilla, graphless conformer throughput (22 QPS) and its accelerated counterpart (117 QPS), representing a five-fold improvement.
This likely arises from the vanilla conformer incurring much kernel launch overhead, on account of its more nested architecture, precisely which CUDA graphs address.


\subsection{Ablation Studies}

\parheader{Data curation}
We measure the quality contribution of each LF, as described in Section \ref{sec:lfs}.
We curate datasets using one additional LF at a time, starting with no LFs, then the session position LF, followed by the ASR confidence LF, and, finally, the rapid repetition LF.
This process results in four datasets for the nested configurations.
To remove transcript diversity and dataset size as confounders, we fix the number of training hours to 200 hours and match the transcript distributions.
We target Wav2vec 2.0-base since it's our deployment model.

We present the ablation results in Table \ref{table:abl-model}.
Each added LF improves the quality, with the first LF having the most impact (1.5 average points for the first vs. 0.1--0.7 for the rest), likely due to diminishing returns.
We note that the ASR confidence score affects CC-20 more than it does CC-LG, possibly because of shorter sessions.

\parheader{Model inference acceleration}
\begin{figure}[t]
    \centering
    \includegraphics[scale=0.43,trim={2.5mm 0 2.2mm 0},clip]{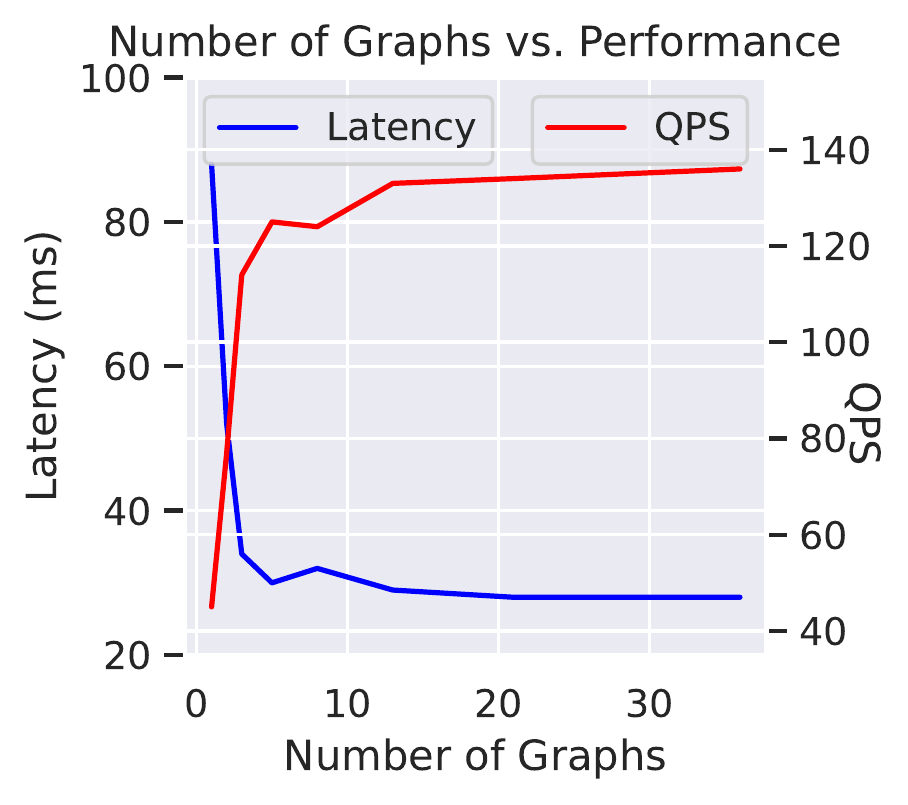}
    \includegraphics[scale=0.43,trim={2.5mm 0 2.2mm 0},clip]{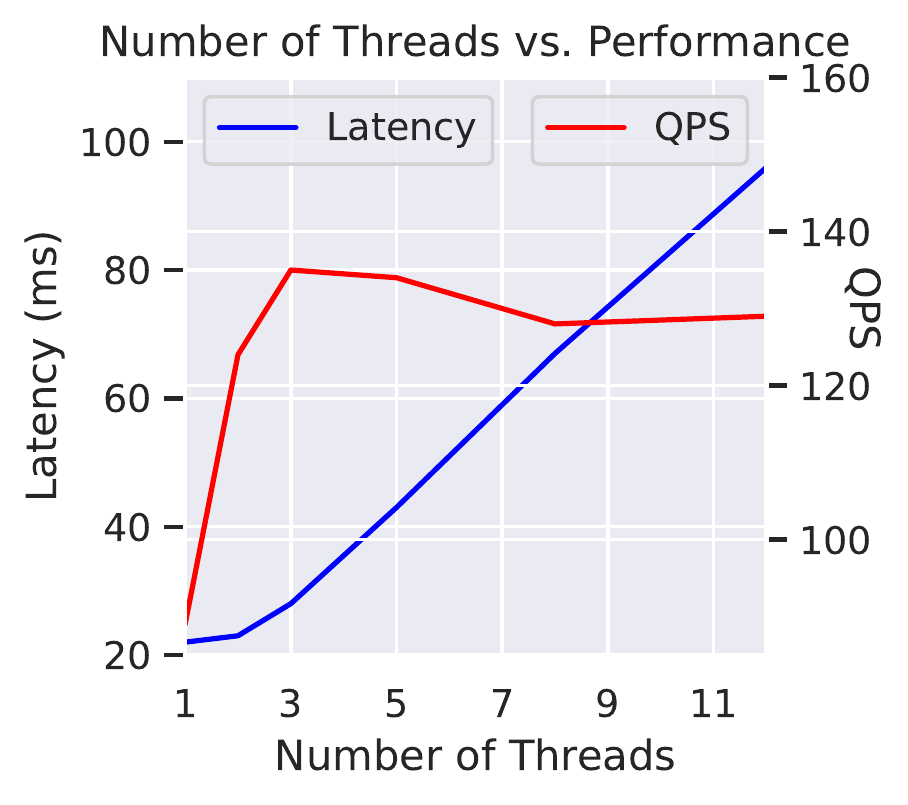}
    \caption{Twin plots of the system latency and throughput plotted against the number of CUDA graphs and inference threads, with the left $y$-axis tracking latency and the right axis throughput.}
    \label{fig:accel}
\end{figure}
We study how the number of CUDA graphs and inference threads (i.e., threads for launching graphs) affects the latency and the throughput, all else being equal.
First, we sweep the number of CUDA graphs and hold the thread count at 3, the optimal value from our experiments.
Next, we vary the thread count and fix the number of graphs at 36, also the best value.
In both settings, we sample 10k queries uniformly at random from production and queue them up in our inference server, which comprises an Nvidia T4 GPU and an eight-core CPU.

We plot our results in Figure~\ref{fig:accel}.
For CUDA graphs, we observe rapidly diminishing returns in both latency and throughput after 5--8 graphs, although they continue to improve until the final value of 36 graphs, the most we can fit in the GPU memory.
For inference threads, we see initially rapid gains in throughput (though not latency) until 4 threads, whereupon throughput tapers slightly and latency grows linearly.
We conjecture that this arises from GPU saturation causing thread contention; while we can certainly push more queries at a time (there being 36 graphs), the GPU can process only 138 queries worth per second.
This results in a backlog of queries when we exceed 3--4 threads, causing linear growth in latency if throughput remains stable.

\subsection{Industrial Considerations}


We deploy SpeechNet as load-balanced Docker Swarm replicas, each exposing a WebSocket API for real-time transcription.
We write the model server in Python and the inference decoder in C++; in particular, we free in the decoder Python's global interpreter lock, a substantial bottleneck in our application.
Our decoder runs faster than all tested open-source CTC decoders do, such as Parlance's \texttt{ctcdecode}, \texttt{pyctcdecode}, and Flashlight.
We execute all graphs in half-precision on separate CUDA streams, further increasing parallelism.

To monitor the reliability of our production system, we measure and expose four key service-level indicators (SLIs): query traffic, server errors, response latency, and system saturation.
Taken together, these represent the so-called ``Google Golden signals,'' a battery of metrics espoused by its namesake.
As is standard in industry, we export real-time metrics to Prometheus, a monitoring system for time series, and then aggregate them in Grafana, a full-stack visualizer.

During the initial release of SpeechNet, these metrics enabled us to detect and mitigate critical imperfections.
In one such case, we observed a large spike in traffic preceding increases in time-out errors and latency.
The spike occurred at the top of the hour, when, due to the nature of television programming, many users issue queries to change shows.
From this evidence, we traced the culprit to our suboptimal decoder implementation, which we promptly fixed.



\section{Related Work}

\parheader{Pretrained ASR models}
Much like natural language processing, the dominant paradigm in the end-to-end speech recognition literature is to pretrain transformers on vast quantities of unlabeled speech and then fine-tune on the labeled datasets.
In their seminal work, \citet{schneider2019wav2vec} pioneer this approach with a contrastive learning objective, calling it Wav2vec.
They further refine it in \citet{baevski2020wav2vec} by introducing discretized representations, naming their model the present Wav2vec 2.0.
Other variants of this model include the Squeezed and Efficient Wav2vec model~\cite{wu2022performance}, which introduces architectural modifications for computational efficiency, and the conformer~\cite{gulati2020conformer}, which adds convolutions in the transformer blocks for better local context modeling.

\parheader{Weakly supervised ASR}
Several papers explore constructing a weakly labeled dataset and training an ASR system with little to no human annotation.
VideoASR~\cite{cheng21weakly} and GigaSpeech~\cite{chen2021gigaspeech} construct speech datasets from videos and subtitles, but this fails in our domain since our users' voice queries differ greatly from those of public sources in both acoustics and text.
For example, our queries contain rare entities (e.g., ``Xfinity Home''), rarely last more than 4--5 seconds, and come from a low-fidelity microphone in frequently noisy households.
Along a separate line, \citet{dufraux2019lead2gold} proposes a label noise-aware objective for ASR; however, this method increases training time by 15--30$\times$, which is too burdensome for us.

\parheader{Model acceleration}
A plethora of model acceleration methods exist for transformers.
In structured pruning, entire blocks of weights are removed, like attention heads~\cite{michel2019sixteen} and weight submatrices~\cite{li2020efficient}, resulting in a more lightweight model.
This comes at the cost of quality, which we can't sacrifice given our thin margin over our third party.
\citet{hinton2015distilling} proposes knowledge distillation, where the outputs of a small model are fine-tuned against those of a large model, but we wish to use the original, pretrained model architecture at runtime for robustness.
Still others propose low bit-width (2--8 bit) quantization~\cite{shen2020q}, which, while quality preserving, has poor conventional GPU software support.
Note that, in this paper, we restricted our experiments to CUDA graph pools because their application does not exclude others.
In fact, when multiple acceleration methods can be applied, \citet{xin2022building} find that the savings are largely cumulative.

\section{Conclusions and Future Work}

In this paper, we explore commercializing a transformer-based, end-to-end speech recognition system without human annotation and with less computational power.
We design three novel labeling functions, derived from implicit user feedback, for Snorkel to construct weakly labeled, in-domain speech datasets from production traffic.
We also propose CUDA graph pools, a novel model acceleration method especially suited for single-example inference, as frequently encountered in production.
Our system, SpeechNet, improves the word-error rate by a relative 8\% and the inference speed by 600\%, compared to our third-party ASR service.
One promising research direction is to extend SpeechNet to the recently released OpenAI Whisper~\cite{radford2022robust}, an ultra large-scale ASR model trained on 680,000 hours of speech, representing the longest corpus to date.


\section*{Limitations}
Our methods primarily apply to companies seeking to build out in-house ASR systems given at least a few thousand customers.
We target business-to-consumer products, not business to business, where clients have wildly different needs without any guarantee on the userbase size (or even existence).
Due to the setting of our work at a for-profit organization, we're also barred from releasing user data and source code out of concerns for privacy and intellectual property.

\bibliography{anthology,custom}

\begin{thebibliography}{25}
\expandafter\ifx\csname natexlab\endcsname\relax\def\natexlab#1{#1}\fi

\bibitem[{Abdou and Scordilis(2004)}]{abdou2004beam}
Sherif Abdou and Michael~S. Scordilis. 2004.
\newblock Beam search pruning in speech recognition using a posterior
  probability-based confidence measure.
\newblock \emph{Speech Communication}.

\bibitem[{Baevski et~al.(2020)Baevski, Zhou, Mohamed, and
  Auli}]{baevski2020wav2vec}
Alexei Baevski, Yuhao Zhou, Abdelrahman Mohamed, and Michael Auli. 2020.
\newblock wav2vec 2.0: A framework for self-supervised learning of speech
  representations.
\newblock \emph{Advances in Neural Information Processing Systems}.

\bibitem[{Beaufays(2022)}]{beaufays2022google}
Fran\c{c}oise Beaufays. 2022.
\newblock Google {C}loud launches new models for more accurate speech {AI}.
\newblock
  https://cloud.google.com/blog/products/ai-machine-learning/google-cloud-updates-speech-api-models-
  for-improved-accuracy.

\bibitem[{Chen et~al.(2021)Chen, Chai, Wang, Du, Zhang, Weng, Su, Povey, Trmal,
  Zhang et~al.}]{chen2021gigaspeech}
Guoguo Chen, Shuzhou Chai, Guanbo Wang, Jiayu Du, Wei-Qiang Zhang, Chao Weng,
  Dan Su, Daniel Povey, Jan Trmal, Junbo Zhang, et~al. 2021.
\newblock Giga{S}peech: An evolving, multi-domain {ASR} corpus with 10,000
  hours of transcribed audio.
\newblock \emph{arXiv:2106.06909}.

\bibitem[{Cheng et~al.(2021)Cheng, Wang, Huang, and Wang}]{cheng21weakly}
Mengli Cheng, Chengyu Wang, Jun Huang, and Xiaobo Wang. 2021.
\newblock Weakly supervised construction of {ASR} systems from massive video
  data.
\newblock In \emph{Proc. Interspeech 2021}.

\bibitem[{Dufraux et~al.(2019)Dufraux, Vincent, Hannun, Brun, and
  Douze}]{dufraux2019lead2gold}
Adrien Dufraux, Emmanuel Vincent, Awni Hannun, Armelle Brun, and Matthijs
  Douze. 2019.
\newblock {Lead2Gold}: Towards exploiting the full potential of noisy
  transcriptions for speech recognition.
\newblock In \emph{2019 IEEE Automatic Speech Recognition and Understanding
  Workshop (ASRU)}.

\bibitem[{Graves(2012)}]{graves2012connectionist}
Alex Graves. 2012.
\newblock Connectionist temporal classification.
\newblock In \emph{Supervised Sequence Labelling with Recurrent Neural
  Networks}. Springer.

\bibitem[{Gray(2019)}]{gray2019getting}
Alan Gray. 2019.
\newblock Getting started with {CUDA} graphs.
\newblock https://developer.nvidia.com/blog/cuda-graphs/.

\bibitem[{Gulati et~al.(2020)Gulati, Qin, Chiu, Parmar, Zhang, Yu, Han, Wang,
  Zhang, Wu et~al.}]{gulati2020conformer}
Anmol Gulati, James Qin, Chung-Cheng Chiu, Niki Parmar, Yu~Zhang, Jiahui Yu,
  Wei Han, Shibo Wang, Zhengdong Zhang, Yonghui Wu, et~al. 2020.
\newblock Conformer: Convolution-augmented transformer for speech recognition.
\newblock \emph{Proc. Interspeech 2020}.

\bibitem[{Hinton et~al.(2015)Hinton, Vinyals, Dean
  et~al.}]{hinton2015distilling}
Geoffrey Hinton, Oriol Vinyals, Jeff Dean, et~al. 2015.
\newblock Distilling the knowledge in a neural network.
\newblock \emph{arXiv:1503.02531}.

\bibitem[{Kuchaiev et~al.(2019)Kuchaiev, Li, Nguyen, Hrinchuk, Leary, Ginsburg,
  Kriman, Beliaev, Lavrukhin, Cook et~al.}]{kuchaiev2019nemo}
Oleksii Kuchaiev, Jason Li, Huyen Nguyen, Oleksii Hrinchuk, Ryan Leary, Boris
  Ginsburg, Samuel Kriman, Stanislav Beliaev, Vitaly Lavrukhin, Jack Cook,
  et~al. 2019.
\newblock Ne{M}o: a toolkit for building ai applications using neural modules.
\newblock \emph{arXiv:1909.09577}.

\bibitem[{Li et~al.(2020)Li, Kong, Zhang, Li, Li, Liu, and
  Ding}]{li2020efficient}
Bingbing Li, Zhenglun Kong, Tianyun Zhang, Ji~Li, Zhengang Li, Hang Liu, and
  Caiwen Ding. 2020.
\newblock Efficient transformer-based large scale language representations
  using hardware-friendly block structured pruning.
\newblock In \emph{Findings of the Association for Computational Linguistics:
  EMNLP 2020}.

\bibitem[{Li and Ture(2020)}]{li2020auto}
Wenyan Li and Ferhan Ture. 2020.
\newblock Auto-annotation for voice-enabled entertainment systems.
\newblock In \emph{Proceedings of the 43rd International ACM SIGIR Conference
  on Research and Development in Information Retrieval}.

\bibitem[{Loshchilov and Hutter(2018)}]{loshchilov2018decoupled}
Ilya Loshchilov and Frank Hutter. 2018.
\newblock Decoupled weight decay regularization.
\newblock In \emph{International Conference on Learning Representations}.

\bibitem[{Michel et~al.(2019)Michel, Levy, and Neubig}]{michel2019sixteen}
Paul Michel, Omer Levy, and Graham Neubig. 2019.
\newblock Are sixteen heads really better than one?
\newblock \emph{Advances in Neural Information Processing Systems}.

\bibitem[{Radford et~al.(2022)Radford, Kim, Xu, Brockman, McLeavey, and
  Sutskever}]{radford2022robust}
Alec Radford, Jong~W. Kim, Tao Xu, Greg Brockman, Christine McLeavey, and Ilya
  Sutskever. 2022.
\newblock Robust speech recognition via large-scale weak supervision.
\newblock \emph{OpenAI Blog}.

\bibitem[{Ratner et~al.(2017)Ratner, Bach, Ehrenberg, Fries, Wu, and
  R{\'e}}]{ratner2017snorkel}
Alexander Ratner, Stephen~H. Bach, Henry Ehrenberg, Jason Fries, Sen Wu, and
  Christopher R{\'e}. 2017.
\newblock Snorkel: Rapid training data creation with weak supervision.
\newblock In \emph{Proceedings of the International Conference on Very Large
  Data Bases}.

\bibitem[{Schneider et~al.(2019)Schneider, Baevski, Collobert, and
  Auli}]{schneider2019wav2vec}
Steffen Schneider, Alexei Baevski, Ronan Collobert, and Michael Auli. 2019.
\newblock {w}av2vec: Unsupervised pre-training for speech recognition.
\newblock \emph{Proc. Interspeech 2019}.

\bibitem[{Shen et~al.(2020)Shen, Dong, Ye, Ma, Yao, Gholami, Mahoney, and
  Keutzer}]{shen2020q}
Sheng Shen, Zhen Dong, Jiayu Ye, Linjian Ma, Zhewei Yao, Amir Gholami,
  Michael~W Mahoney, and Kurt Keutzer. 2020.
\newblock Q-{BERT}: Hessian based ultra low precision quantization of {BERT}.
\newblock In \emph{Proceedings of the AAAI Conference on Artificial
  Intelligence}.

\bibitem[{Tang et~al.(2019{\natexlab{a}})Tang, Lu, Liu, Mou, Vechtomova, and
  Lin}]{tang2019distilling}
Raphael Tang, Yao Lu, Linqing Liu, Lili Mou, Olga Vechtomova, and Jimmy Lin.
  2019{\natexlab{a}}.
\newblock Distilling task-specific knowledge from {BERT} into simple neural
  networks.
\newblock \emph{arXiv:1903.12136}.

\bibitem[{Tang et~al.(2019{\natexlab{b}})Tang, Ture, and Lin}]{tang2019yelling}
Raphael Tang, Ferhan Ture, and Jimmy Lin. 2019{\natexlab{b}}.
\newblock Yelling at your {TV}: An analysis of speech recognition errors and
  subsequent user behavior on entertainment systems.
\newblock In \emph{Proceedings of the 42nd International ACM SIGIR Conference
  on Research and Development in Information Retrieval}.

\bibitem[{Vaswani et~al.(2017)Vaswani, Shazeer, Parmar, Uszkoreit, Jones,
  Gomez, Kaiser, and Polosukhin}]{vaswani2017attention}
Ashish Vaswani, Noam Shazeer, Niki Parmar, Jakob Uszkoreit, Llion Jones,
  Aidan~N. Gomez, {\L}ukasz Kaiser, and Illia Polosukhin. 2017.
\newblock Attention is all you need.
\newblock \emph{Advances in Neural Information Processing Systems}.

\bibitem[{Wolf et~al.(2019)Wolf, Debut, Sanh, Chaumond, Delangue, Moi, Cistac,
  Rault, Louf, Funtowicz et~al.}]{wolf2019huggingface}
Thomas Wolf, Lysandre Debut, Victor Sanh, Julien Chaumond, Clement Delangue,
  Anthony Moi, Pierric Cistac, Tim Rault, R{\'e}mi Louf, Morgan Funtowicz,
  et~al. 2019.
\newblock Hugging{F}ace's {T}ransformers: State-of-the-art natural language
  processing.
\newblock \emph{arXiv:1910.03771}.

\bibitem[{Wu et~al.(2022)Wu, Kim, Pan, Han, Weinberger, and
  Artzi}]{wu2022performance}
Felix Wu, Kwangyoun Kim, Jing Pan, Kyu~J. Han, Kilian~Q. Weinberger, and Yoav
  Artzi. 2022.
\newblock Performance-efficiency trade-offs in unsupervised pre-training for
  speech recognition.
\newblock In \emph{IEEE International Conference on Acoustics, Speech and
  Signal Processing (ICASSP)}.

\bibitem[{Xin et~al.(2022)Xin, Tang, Jiang, Yu, and Lin}]{xin2022building}
Ji~Xin, Raphael Tang, Zhiying Jiang, Yaoliang Yu, and Jimmy Lin. 2022.
\newblock Building an efficiency pipeline: Commutativity and cumulativeness of
  efficiency operators for transformers.
\newblock \emph{arXiv:2208.00483}.

\end{thebibliography}
\bibliographystyle{acl_natbib}

\appendix

\section{Computational Environment}
We train all models on Amazon \texttt{p3.2xlarge} instances running HuggingFace Transformers 4.15.0, from which we borrow the SEW and Wav2vec implementations; PyTorch 1.11.0 (CUDA 10.2), a popular deep learning framework; Nvidia's NeMo library, which we depend on for the Conformer implementation; and SentencePiece 0.1.94, which we use for the character-based tokenizer.
We implement our CTC decoder in C++14, interfacing with Python using pybind11 and the development libraries for SentencePiece and PyTorch (LibTorch).
We serve users on geographically dispersed data centers on the American east and west coasts, running Nginx-load-balanced boxes with Nvidia T4s.

\section{Dataset and Production Statistics}

We curated CC-20 sampled across weeks of traffic, with training, dev, and test coming from separate speakers.
We constructed CC-LG's training set sampled from 2 days of traffic between July 3 and July 5, 2022 and the development/test sets from separate users sampled a day after the training set.

\begin{figure}[h]
    \centering
    \includegraphics[scale=0.108,clip,trim=0.5cm 1cm 0.5cm 0]{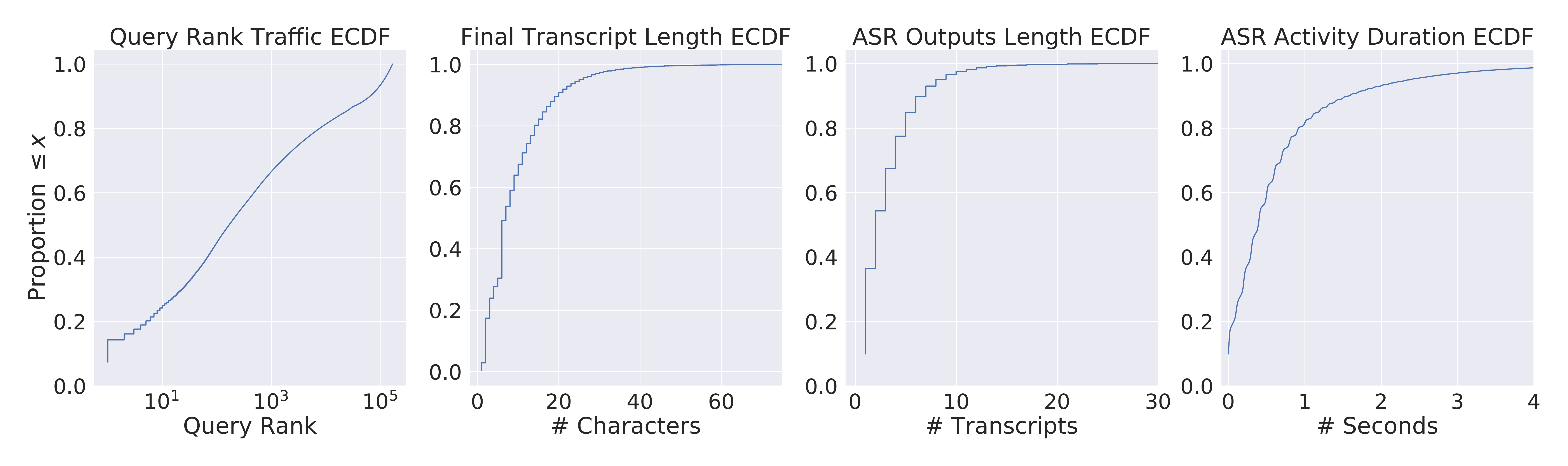}
    \caption{Distribution statistics of our in-production queries.}
    \label{fig:asr-stats}
\end{figure}

We present detailed production statistics of our queries in Figure~\ref{fig:asr-stats}.
The query distributions have large right skew, with the top 1000 queries making up nearly 70\% of the traffic, as the first subplot shows.
Our queries are lexically simple, e.g., ``Hulu,'' ``Free movies for me,'' etc., as the second subgraph shows.
The third and fourth subgraphs denote the activity of the ASR system---most queries are less than 1--2 seconds in speech (not necessarily total audio length).

\section{Training Details}

For all models, on CC-LG, we first resize and re-initialize the final linear layer to match our vocabulary size, then fine-tune just the output linear layer (as recommended in the original Wav2vec paper) for 30k steps.
Next, we ran 750k optimization steps on the ``raw'' training set.
Then, we train for an additional 100k steps on the ``weak'' subset, if applicable.
If it's the raw training run, we still train for an additional 100k steps, but on the ``raw'' training set as usual.
That is, all configurations on CC-20 use 850k optimization steps.
On CC-20, we use 10k steps for the initial output layer fine-tuning and then ran 50k optimization steps for all models.
We use the AdamW~\cite{loshchilov2018decoupled} optimizer with a batch size of 8 for all runs.
We decode all model outputs using a beam size of 15 and a beam cutoff of 30.
All model weights are initialized from the respective model cards on HuggingFace's model zoo.
We describe model-specific hyperparameters:

\parheader{SEW}
We optimize our models using a learning rate of $3 \times 10^{-6}$, determined from preliminary experiments across several choices spanning different orders of magnitude.
SEW operates on the raw audio waveform.

\parheader{Wav2vec 2.0}
We use a learning rate of $2 \times 10^{-6}$, determined similarly.
Wav2vec 2.0 operates on the raw audio waveform as well.

\parheader{Conformer}
As is standard, we transform all audio amplitudes to 80-dimensional Mel spectrograms before being input to the Conformer encoder.
We pick a learning rate of $5 \times 10^{-6}$ using the same procedure as the other models do.

\end{document}